\documentclass[journal]{IEEEtran}

\usepackage[utf8]{inputenc} % allow utf-8 input
\usepackage[T1]{fontenc}    % use 8-bit T1 fonts
\usepackage{hyperref}       % hyperlinks
\usepackage{url}            % simple URL typesetting
\usepackage{booktabs}       % professional-quality tables
\usepackage{amsfonts}       % blackboard math symbols
\usepackage{nicefrac}       % compact symbols for 1/2, etc.
\usepackage{microtype}      % microtypography
\usepackage{graphicx}      % resizebox, inter alia
\usepackage{enumitem}      % list control
\usepackage{siunitx}       % number columns in tables etc
\usepackage{etoolbox}
\robustify{\bfseries}
\usepackage{subcaption}
\usepackage{float}

\usepackage{arydshln}
\usepackage{amsmath}
\usepackage{multirow, multicol}

\newcommand{\reportvalbrief}[2]{#1}

\title{Overview of the Ninth Dialog System\\ Technology Challenge: DSTC9}

\author{
  \bf{Chulaka Gunasekara, Seokhwan Kim, Luis Fernando D'Haro, Abhinav Rastogi, Yun-Nung Chen, }\\
  \bf{Mihail Eric, Behnam Hedayatnia, Karthik Gopalakrishnan, Yang Liu, Chao-Wei Huang,}\\
  \bf{ Dilek Hakkani-T\"ur, Jinchao Li, Qi Zhu, Lingxiao Luo, Lars Liden, Kaili Huang, Shahin Shayandeh, Runze Liang, Baolin Peng, Zheng Zhang, Swadheen Shukla, Minlie Huang, Jianfeng Gao,}\\
  \bf{Shikib Mehri, Yulan Feng, Carla Gordon, Seyed Hossein Alavi, David Traum, Maxine Eskenazi,}\\
  \bf{Ahmad Beirami, Eunjoon (EJ) Cho, Paul A. Crook, Ankita De, Alborz Geramifard, Satwik Kottur, Seungwhan Moon, Shivani Poddar, Rajen Subba\thanks{Every author has equal contribution}}
}

\begin{document}

\maketitle

\begin{abstract}
  This paper introduces the Ninth Dialog System Technology Challenge (DSTC-9).
  This edition of the DSTC focuses on applying end-to-end dialog technologies for four distinct tasks in dialog systems, namely, 1. Task-oriented dialog Modeling with unstructured knowledge access, 2. Multi-domain task-oriented dialog, 3. Interactive evaluation of dialog, and 4. Situated interactive multi-modal dialog.
  This paper describes the task definition, provided datasets, baselines and evaluation set-up for each track.
  We also summarize the results of the submitted systems to highlight the overall trends of the state-of-the-art technologies for the tasks.
\end{abstract}

\section{Introduction}
The Dialog System Technology Challenge (DSTC) is a one of the leading series of research competitions in the space of dialog systems. Since the inception in 2013, DSTC has been accelerating the development of dialog technologies, by bringing the leading researchers and engineers together to solve important problems in dialog systems. The challenge has been evolving every year to cater the demand and the interest of the dialog community to foster the development of technology.

The first Dialog System Technology Challenge \cite{williams2013dialog} used human-to-bot dialogs in the bus timetable domain. Dialog State Tracking Challenges 2 \cite{henderson2014second} and 3 \cite{henderson2014third} used restaurant reservation application which introduced more complicated and dynamic dialog states.  Dialog State Tracking Challenge 4 \cite{kim2017fourth} and Dialog State Tracking Challenge 5 \cite{kim2016fifth} moved to tracking human-to-human dialogs in mono and cross-language settings. From the sixth challenge~\cite{hori2018overview}, the DSTC has rebranded itself as ``Dialog System Technology Challenge'' and organized multiple tracks in parallel to address a wider variety of dialog related problems. The tracks in DSTC-6 were focused on end-to-end conversation modeling and dialog breakdown detection. DSTC-7~\cite{yoshino2019dialog} focused on developing end-to-end dialog technologies for noetic response selection~\cite{dstc7task1,gunasekara2019dstc7}, grounded response generation~\cite{galley2019grounded}, and audio visual scene aware dialog~\cite{alamri2018audio}. More recently in DSTC-8 \cite{kim2019eighth} the focus has been on diverse set of four tracks including, multi-domain task completion, predicting responses, audio visual scene-aware dialog and schema-guided dialog state tracking. 

For the ninth edition of the DSTC, we received nine track proposals from the leading research organizations and top universities. The proposals went through a formal peer review process focusing on each task's potential for, (a) impact to the community, (b) novelty of the task, (c) feasibility of the proposal, and (d) potential participants. The DSTC-8 participants were also asked to provide their feedback on the presented track proposals through a survey, and the responses were also considered in the evaluation. Finally, we ended up with the four main tracks including three newly introduced tasks and one follow-up task from DSTC-8.

The track, \textit{Beyond Domain APIs: Task-oriented Conversational Modeling with Unstructured Knowledge Access} (Track 1), aims to support frictionless task-oriented scenarios, where the flow of the conversation does not break when users have requests that are out of the scope of APIs/DB but potentially are already available in external knowledge sources. Track 2, \textit{Multi-domain Task-oriented Dialog Challenge II}, is a continuation of last year, and focuses on end-to-end multi-domain task completion dialog and cross-lingual multi-domain dialog state tracking. The track 3 of this year, \textit{Interactive Evaluation of Dialog}, aims to take the first step in expanding dialog research beyond datasets and challenges the participants to develop dialog systems that can converse effectively in interactive environments with real users. \textit{SIMMC: Situated Interactive Multi-Modal Conversational AI} (track 4) is aimed at laying the foundations for the real-world assistant agents that can handle multi-modal inputs, and perform multi-modal actions. 

The following sections describe the details of each track. 

\section{Track 1 - Beyond Domain APIs - Task-Oriented Conversational Modeling with Unstructured Knowledge Access}
\subsection{Track Overview}
Most prior work on task-oriented dialog systems has been restricted to a limited coverage of domain APIs. However, users often have domain related requests that are not covered by the APIs.
This challenge track aims to expand the coverage of task-oriented dialog systems by incorporating external unstructured knowledge sources. 
There are three main tasks in this track as introduced in~\cite{kim-etal-2020-beyond}: knowledge-seeking turn detection, knowledge selection, and knowledge-grounded response generation (Table~\ref{tbl:track1_tasks}).

\begin{table}[t]
  \centering
  \caption{Summary of Track 1 tasks}
  \small
  \begin{tabular}{p{1.2cm} p{6.8cm}}
    \hline
    \textbf{Task \#1} & \textbf{Knowledge-seeking Turn Detection}\\ \hdashline[.4pt/1pt] \hdashline[.4pt/1pt]
    Goal & To decide whether to continue existing flow or trigger the knowledge access branch for a given utterance and dialog history\\
    Input & Current user utterance, dialog context, and domain API and knowledge sources\\
    Output & Binary class (requires knowledge access or not)\\ \hline
    \textbf{Task \#2} & \textbf{Knowledge Selection}\\ \hdashline[.4pt/1pt]
    Goal & To select proper knowledge sources from the domain knowledge-base given dialog context at each turn with knowledge access\\
    Input & Current user utterance, dialog context, and the entire set of knowledge candidates\\
    Output & Ranking of top-$k$ knowledge candidates\\ \hline
    \textbf{Task \#3} & \textbf{Knowledge-grounded Response Generation}\\ \hdashline[.4pt/1pt]
    Goal & To generate a system response for a given triple of input utterance, dialog context, and the selected knowledge sources\\
    Input & Current user utterance, dialog context, and selected knowledge sources\\
    Output & Generated system response\\ \hline
  \end{tabular}
  \label{tbl:track1_tasks}
\end{table}

\subsection{Data}
This challenge track uses two different data sets (Table~\ref{tbl:data_stats}).
The first data is an augmented version of MultiWOZ 2.1~\cite{eric2019multiwoz21} that includes newly introduced knowledge-seeking turns in the MultiWOZ conversations.
The data augmentation was incrementally done by the crowdsourcing tasks described in~\cite{kim-etal-2020-beyond}.
A total of 22,834 utterance pairs were newly collected based on 2,900 knowledge candidates from the FAQ webpages about the domains and the entities in MultiWOZ databases.
For the challenge track, we divided the whole data into three subsets: train, validation and test.
The first two sets were released in the development phase along with the ground-truth annotations and human responses for participants to develop their models.

In the evaluation phase, we released the test split of the augmented MultiWOZ 2.1 and the other conversations collected from scratch about touristic information for San Francisco.
To evaluate the generalizability of models, the new conversations cover knowledge, locale and domains that are unseen from the train and validation data sets.
In addition, this test set includes not only written conversations, but also spoken dialogs to evaluate system performance across different modalities~\cite{gopalakrishnan2020neural}.
All the backend resources for this data collection were also released, which includes 9,139 knowledge snippets and 855 database entries for San Francisco.

\begin{table}[t]
  \centering
  \caption{Statistics of the Track 1 data sets}
  \small
  \begin{tabular}{l l r r r}
  \hline
    & & \# & total \# & \# knowledge \\
    Source & Split & dialogs & instances & seeking turns \\ \hline
    \multirow{3}{*}{MultiWOZ} & Train & 7,190 & 71,348 & 19,184 \\
    & Valid & 1,000 & 9,663 & 2,673 \\
    & Test & 977 & 2,084 & 977 \\ \hdashline[.4pt/1pt]
    \multirow{2}{*}{SF} & Written & 900 & 1,834 & 900\\
    & Spoken & 107 & 263 & 104\\ \hline
    % Total & 10,438 & 9,072 & 161,192
  \end{tabular}
  \label{tbl:data_stats}
\end{table}

\subsection{Evaluation Criteria}

\begin{table}[t]
  \centering
  \caption{Objective evaluation metrics for the Track 1 tasks}
  \small
  \begin{tabular}{l l}
  \hline
    Task & Metrics \\ \hline
    Task \#1 & Precision/Recall/F-measure \\\hdashline[.4pt/1pt]
    Task \#2 & MRR@5, Recall@1, Recall@5 \\ \hdashline[.4pt/1pt]
    Task \#3 & BLEU-1, BLEU-2, BLEU-3, BLEU-4, METEOR\\
    & ROUGE-1, ROUGE-2, ROUGE-L \\ \hline
  \end{tabular}
  \label{tbl:track1_metrics}
\end{table}

Each participating team submitted up to five system outputs each of which contains the results for all three tasks on the unlabeled test instances.
We first evaluated each submission using the task-specific objective metrics (Table~\ref{tbl:track1_metrics}) by comparing to the ground-truth labels and responses.
Considering the dependencies between the tasks in the pipelined architecture, the final scores for knowledge selection and knowledge-grounded response generation are computed by considering the first step  knowledge-seeking turn detection recall and precision performance, as follows:
\begin{equation*}
  f(x) = \left\{ \begin{array}{ll}
    1 & \mbox{if $x$ is a knowledge-seeking turn,}\\
    0 & \mbox{otherwise}
  \end{array} \right.
\end{equation*}
\begin{equation*}
  \tilde{f}(x) = \left\{ \begin{array}{ll}
    1 & \mbox{if $x$ is predicted as a knowledge-seeking turn,}\\
    0 & \mbox{otherwise}
  \end{array} \right.
\end{equation*}

\begin{equation*}
  S_p(X) = \frac{\sum_{x_i \in X}\left(s(x_i) \cdot f(x_i) \cdot \tilde{f}(x_i)\right)}{\sum_{x_i \in X}\tilde{f}(x_i)},
\end{equation*}
\begin{equation*}
  S_r(X) = \frac{\sum_{x_i \in X}\left(s(x_i) \cdot f(x_i) \cdot \tilde{f}(x_i)\right)}{\sum_{x_i \in X}f(x_i)},
\end{equation*}
\begin{equation}
  S_f(X) = \frac{2 \cdot S_p(X) \cdot S_r(X)}{S_p(X) + S_r(X)},
  \label{eq:end_to_end_score}
\end{equation}
where $s(x)$ is the knowledge selection or response generation score in a target metric for a single instance $x \in X$.

Then, we aggregated a set of multiple scores across different tasks and metrics into a single overall score computed by the mean reciprocal rank, as follows:
\begin{equation}
  S_{overall}(e) = \frac{1}{|M|}\sum_{i=1}^{|M|}\frac{1}{rank_i(e)},
  \label{eq:overall_score}
\end{equation}
where $rank_i(e)$ is the ranking of the submitted entry $e$ in the $i$-th metric against all the other submissions and $M$ is the number of metrics we considered. 

Based on the overall objective score, we selected the finalists to be manually evaluated by the following two crowd sourcing tasks:
\begin{itemize}
\item Appropriateness: This task asks crowd workers to score how well a system output is naturally connected to a given conversation on a scale of 1-5.
\item Accuracy: This task asks crowd workers to score the accuracy of a system output based on the provided reference knowledge on a scale of 1-5.
\end{itemize}
In both tasks, we assigned each instance to three crowd workers and took their average as the final human evaluation score for the instance.
Those scores were then aggregated over the entire test set following Equation~\ref{eq:end_to_end_score}, i.e., weighted by the knowledge-seeking turn detection performance.
Finally, we used the average of the Appropriateness and Accuracy scores to determine the official ranking of the systems in the challenge track.

\subsection{Results}
We received 105 entries in total submitted from 24 participating teams.
To preserve anonymity, the teams were identified by numbers from 1 to 24, while our baseline~\cite{kim-etal-2020-beyond} was marked as team 0.
Table~\ref{tbl:track1_objective} shows the objective evaluation results of the best entry from each team in the featured metrics.
The full scores with all the submitted entries and the other metrics are available on the track repository\footnote{\url{https://github.com/alexa/alexa-with-dstc9-track1-dataset}}.
Most entries outperformed the baseline in all three tasks.
In particular, the best entry from Team 3 achieved over 99\% F-measure for knowledge-seeking turn detection, and also the highest scores in the BLEU and ROUGE variants for the response generation task.
On the other hand, Team 19 was the best in the knowledge selection metrics and Team 15 was better than all the other teams in METEOR for generation.
We calculated the overall score (Equation~\ref{eq:overall_score}) of each entry and selected 12 finalists,  corresponding to the best entry from each of the top 12 teams.

\begin{table}[t]
  \caption{Objective evaluation results of the best entry from each team in the featured metrics for the Track 1 tasks. Team 0 is the baseline. Bold denotes the best result in each column and $^*$ indicates the finalists.}
  \label{tbl:track1_objective}
  \centering
  \small
  \begin{tabular}{l r r r r r}
  \hline
    & Task\#1 & Task\#2 & \multicolumn{3}{c}{Task\#3}\\
    Team & F1 & R@1 & BLEU-1 & METEOR & ROUGE-L \\ \hline
    0 & 0.9455 & 0.6201 & 0.3031 & 0.2983 & 0.3039 \\ \hdashline[.4pt/1pt]
    1 & 0.9721 & 0.8255 & 0.3368 & 0.3342 & 0.3364 \\
    2 & 0.9644 & 0.8584 & 0.3338 & 0.3322 & 0.3330 \\
    3$^*$ & \bf{0.9911} & 0.9013 & \bf{0.3879} & 0.3914 & \bf{0.3885} \\
    4 & 0.8998 & 0.6950 & 0.3025 & 0.3001 & 0.2990 \\
    5 & 0.9428 & 0.7055 & 0.3218 & 0.3266 & 0.3174 \\
    6 & 0.9838 & 0.8531 & 0.3371 & 0.3341 & 0.3362 \\
    7$^*$ & 0.9702 & 0.8988 & 0.3752 & 0.3854 & 0.3702 \\
    8 & 0.9530 & 0.7403 & 0.3135 & 0.3097 & 0.3119 \\
    9 & 0.9242 & 0.7931 & 0.3154 & 0.3159 & 0.3101 \\
    10$^*$ & 0.9726 & 0.9158 & 0.3684 & 0.3719 & 0.3692 \\
    11$^*$ & 0.9675 & 0.8702 & 0.3743 & 0.3854 & 0.3797 \\
    12 & 0.9540 & 0.8011 & 0.3411 & 0.3526 & 0.3401 \\
    13$^*$ & 0.9819 & 0.8434 & 0.3787 & 0.3902 & 0.3619 \\
    14 & 0.9252 & 0.6466 & 0.3019 & 0.2974 & 0.3003 \\
    15$^*$ & 0.9803 & 0.8975 & 0.3779 & \bf{0.3931} & 0.3765 \\
    16 & 0.9549 & 0.7327 & 0.3351 & 0.3334 & 0.3364 \\
    17$^*$ & 0.9839 & 0.8734 & 0.3699 & 0.3724 & 0.3687 \\
    18$^*$ & 0.9635 & 0.8994 & 0.3806 & 0.3864 & 0.3726 \\
    19$^*$ & 0.9886 & \bf{0.9235} & 0.3803 & 0.3869 & 0.3738 \\
    20$^*$ & 0.9711 & 0.8628 & 0.3619 & 0.3637 & 0.3535 \\
    21$^*$ & 0.9396 & 0.8269 & 0.3551 & 0.3594 & 0.3576 \\
    22 & 0.7803 & 0.2831 & 0.1792 & 0.1583 & 0.1852 \\
    23$^*$ & 0.9618 & 0.8959 & 0.3534 & 0.3565 & 0.3519 \\
    24 & 0.3471 & 0.0017 & 0.0835 & 0.0796 & 0.0866 \\
    \hline
  \end{tabular}
\end{table}

Table~\ref{tbl:track1_human} shows the final ranking of the Track 1 participating teams based on the human evaluation scores of the finalist entries.
The top three teams (Team 19, 3 and 10) commonly used ensemble of large-scale pre-trained language models in their best entries.
Team 19 won the challenge track with the highest scores for both Accuracy and Appropriateness, most likely because of their better performance in the knowledge selection task (as in the objective evaluation results).
To compare the importance of each task towards end-to-end performance, we calculated the Spearman's rank correlation coefficient of the ranked lists of all the entries in every pair of objective and human evaluation metrics.
As a result, Recall@1 for the knowledge selection task shows a strong correlation with the averaged human evaluation ranking at 0.8601, which is significantly higher than 0.7692 and 0.6503 with F-measure for the knowledge-seeking turn detection and BLEU-1 for the response generation, respectively.
This implies that the knowledge-selection is a key task to improve end-to-end performance.

\begin{table}[t]
  \caption{Human evaluation results for the Track 1 finalists}
  \label{tbl:track1_human}
  \centering
  \small
  \begin{tabular}{l l l r r r}
  \hline
    Rank & Team & Entry & Accuracy & Appropriateness & Average \\ \hline
    & \multicolumn{2}{l}{Ground-truth} & 4.5930 & 4.4513 & 4.5221  \\ \hdashline[.4pt/1pt]
    1 & 19 & 2 & \bf{4.3917} & \bf{4.3922} & \bf{4.3920}  \\
    2 & 3 & 1 & 4.3480 & 4.3634 & 4.3557  \\
    3 & 10 & 0 & 4.3544 & 4.3201 & 4.3373  \\
    4 & 15 & 3 & 4.3793 & 4.2755 & 4.3274  \\
    5 & 17 & 0 & 4.3360 & 4.3076 & 4.3218  \\
    6 & 7 & 4 & 4.3308 & 4.2989 & 4.3149  \\
    7 & 18 & 3 & 4.3309 & 4.2859 & 4.3084  \\
    8 & 13 & 3 & 4.3763 & 4.2360 & 4.3061  \\
    9 & 23 & 0 & 4.3082 & 4.2665 & 4.2874  \\
    10 & 11 & 3 & 4.2722 & 4.2619 & 4.2670  \\
    11 & 20 & 4 & 4.2283 & 4.2486 & 4.2384  \\
    12 & 21 & 3 & 4.1060 & 4.1560 & 4.1310  \\ \hdashline[.4pt/1pt]
    & \multicolumn{2}{l}{Baseline} & 3.7155 & 3.9386 & 3.8271  \\
    \hline
  \end{tabular}
\end{table}

\section{Track 2 - Multi-domain Task-oriented Dialog Challenge II}
We provide two tasks in the multi-domain task-oriented dialog setting. One is the end-to-end task-oriented dialog task aiming to solve the complexity of building end-to-end dialog systems. The other is cross-lingual dialog state tracking (DST) to address the language adaption problem for the DST task.

\subsection{End-to-end Task-oriented Dialog Task}
 This task is a continuation of last year at DSTC8 \cite{DBLP:journals/corr/abs-1911-06394}. Participants will develop an end-to-end task-oriented dialog system that takes natural language as input and generates natural language response as output in the travel planning setting. Both the evaluation result of last year's challenge \cite{li2020results} and empirical analysis of models in ConvLab \cite{takanobu-etal-2020-goal} show that the best rule-based pipeline systems outperform systems assembled using state-of-the-art component-wise machine learning models. From the results, we've also observed a discrepancy between the performance of component-wise models using corpus-based evaluation and that of the entire system using the end-to-end evaluation. These findings are consistent with the landscape of dialog development technology stacks in the industry. However, interestingly, the winning team built their model based on GPT-2 \cite{ham2020end}, and achieved significant improvement over other teams with regards to success rate, understanding score, and response score at the human evaluation phase. Meanwhile, by using similar model training paradigms, SOLOIST \cite{peng2020soloist} and SimpleTOD \cite{hosseini2020simple} shortly achieved top performance in the MultiWOZ leaderboard by leveraging GPT-2 \cite{radford2019language}.
 
This year, we continue with the end-to-end task-oriented dialog task, aiming to promote the technology of building end-to-end dialog systems one step further. Like last year, participants are encouraged to explore all possible approaches, and there is no restriction on dialog system architecture.

\subsubsection{Data}
Participants are expected to build dialog systems based on MultiWOZ 2.1 \cite{eric2019multiwoz21}, a multi-domain dialog dataset spanning 7 distinct domains containing over 10,000 dialogs under the travel planning setting. Compared with MultiWOZ 2.0 \cite{DBLP:conf/emnlp/BudzianowskiWTC18}, MultiWOZ 2.1 re-annotated states to fix the noisy annotation and incorporated user dialog act annotation. Although the dialog system is evaluated under MultiWOZ 2.1, participants can leverage any public datasets, pre-trained models, or other resources to build the dialog system.

\subsubsection{Evaluation Criteria}
ConvLab-2 \cite{DBLP:conf/acl/ZhuZFLTLPGZH20} is employed as the platform for dialog development and evaluation. As the successor of ConvLab \cite{DBLP:conf/acl/LeeZTZZLLPLHG19}, ConvLab-2 provides a user simulator and evaluator for MultiWOZ 2.1 so that the participants can effectively run offline experiments and evaluations. Specifically, we offer two evaluation approaches:
\paragraph{Automatic Evaluation} The dialog system is evaluated via conversing with an end-to-end user simulator. The simulator is constructed by assembling a BERT-based natural language understanding model \cite{DBLP:conf/naacl/DevlinCLT19} , an agenda-based user simulator \cite{schatzmann2007agenda} and a rule-based natural language generation module. A dialog is successful only if all requested slots are filled with grounded values in the database, and the booking is successful. We report metrics including success rate, book rate, number of turns for each dialog, and precision/recall/F1 score for slot prediction. 
\paragraph{Human Evaluation} In human evaluation, Amazon Mechanic Turkers communicate with the dialog systems via natural language, judge whether the dialog is successful, and provide scores based on language understanding correctness and response appropriateness on a 5 point Likert-scale. Since MTurkers do not directly access the back-end database, we also report the success rate with grounding after verifying whether the requested slot values returned by the dialog systems match the database record. We take the average value of success rate with grounding and without grounding as the final ranking.

\subsubsection{Results}
As per our submission policy, each team is allowed to submit up to 5 models. We received 34 models in total from  10 teams. Table \ref{tab:task1:automatic} lists the automatic evaluation result for the best models of each team. We filtered out low-performance models based on the automatic evaluation result while keeping the best model for each team and sent the remaining models for human evaluation. With this process, 21 models were evaluated in human evaluation, with the performance of each team's best model listed in Table \ref{tab:task1:human_eval}. 

Team 1 achieves the top 1 performance in both automatic and human evaluation by constructing an end-to-end dialog system with the pre-trained dialog generation model PLATO-2 \cite{bao2020plato}. This model generates the dialog state, system action, and system response simultaneously, given the dialog context. The dialog state is used as the constraint for database query, and the system action is then refreshed according to the queried results to re-generate the final system response. Team 2 achieves the same ranking as Team 1 in the human evaluation using a similar hybrid end-to-end neural model. It borrows idea from \cite{peng2020soloist} and \cite{DBLP:conf/acl/GururanganMSLBD20}, uses GPT-2 as the backend for pre-training and fine-tuning and add various pre/post-processing modules to improve model generalization ability. An additional fault tolerance mechanism is also added to correct errors.

\sisetup{table-column-width=11mm,round-precision=3,tight-spacing=false,table-format=1.3}
\newcolumntype{C}{>{\centering\arraybackslash}X}
\begin{table}
  %  \centering
  \caption{Automatic Evaluation Result (Best Submissions) }
    \small
        \centering
      %   \begin{tabular}{@{}cS@{\;}Scccc@{}}
      \begin{tabular}{@{}cccccc@{}}
        \toprule
           Team & SR & CR & BR &  Inform P/R/F1  & Turn S/A\\
        \midrule

1         & {\bf{93}}          & 95.2         & 94.6     & 84.1/96.2/88.1 & 12.5/12.7    \\
2          & 91.4        & 96.9         & 96.2     & 80.2/97.3/86.0 & 15.3/15.7    \\
3         & 90.8        & 94.4         & 96.7     & 81.0/95.4/85.9 & 13.4/13.6    \\
4       & 89.8        & 94.6         & 96.3     & 72.4/96.0/80.1 & 15.1/15.8    \\
5         & 83.3        & 88.5         & 89.1     & 81.1/90.3/83.5 & 13.5/13.8    \\
6        & 67.7        & 88.5         & 90.8     & 70.4/85.6/75.2 & 12.8/14.2    \\
7          & 57.8        & 87.1         & 85       & 68.7/81.6/72.6 & 13.7/16.4    \\
8         & 52.6        & 66.9         & 66.7     & 57.5/80.7/64.8 & 13.2/22.5    \\
9        & 44.4        & 50           & 26.5     & 57.9/64.5/58.9 & 12.2/14.6    \\
10        & 21.4        & 40.7         & 0        & 55.4/60.0/54.1 & 11.0/25.9   \\
Baseline & 85 & 92.4 & 91.4 & 79.3/94.9/84.5 & 13.8/14.9 \\
\bottomrule\\[-3.5mm]
\multicolumn{6}{@{}l@{}}{\scriptsize SR: Success Rate, CR: Complete Rate, BR: Book Rate,  Inform P/R/F1: Prec./Recall/F1} \\[-1mm]
% \multicolumn{6}{@{}l@{}}{\scriptsize \tiny Inform P/R/F1: Precision/Recall/F-1 score of slots prediction,}  \\[-1mm]
\multicolumn{6}{@{}l@{}}{\scriptsize score of slots prediction, Turn S/A: Turns for successful and all dialogs, respectively.}
        \end{tabular}
    \label{tab:task1:automatic}
\end{table}

\newcolumntype{C}{>{\centering\arraybackslash}X}
\sisetup{table-column-width=11mm,round-precision=3,tight-spacing=false,table-format=1.3}
\begin{table}
    \footnotesize
    \caption{Human Evaluation Result (Best Submissions)}
        \centering
        % \begin{tabularx}{\linewidth}{@{}Cccccccc@{}}
        \begin{tabular}{@{}cccccccc@{}}
        \toprule
           Team & SRa & SRwg & SRog & Under. & Appr. & Turn & Rank \\
        \midrule

 1       & {\bf{74.8}} & 70.2 & 79.4 & 4.54 & 4.47 & 18.5 & 1  \\
2        & {\bf{74.8}} & 68.8 & 80.8 & 4.51 & 4.45 & 19.4 & 1  \\
 7        & 72.3 & 62   & 82.6 & 4.53 & 4.41 & 17.1 & 3  \\
6        & 70.6 & 60.8 & 80.4 & 4.41 & 4.41 & 20.1 & 4  \\
3         & 67.8 & 60   & 75.6 & 4.56 & 4.42 & 21  & 5   \\
4         & 60.3 & 51.4 & 69.2 & 4.49 & 4.22 & 17.7 & 6   \\
5         & 58.4 & 50.4 & 66.4 & 4.15 & 4.06 & 19.7 & 7 \\
9        & 55.2 & 43.2 & 67.2 & 4.15 & 3.98 & 19.2 & 8  \\
 8       & 35   & 26   & 44   & 3.27 & 3.15 & 18.5 & 9  \\
 10      & 19.5 & 6    & 33   & 3.23 & 2.93 & 18.8 & 10  \\
Baseline   & 69.6 & 56.8 & 82.4 & 4.34 & 4.18 & 18.5 & N/A \\
\bottomrule \\[-3.5mm]
\multicolumn{8}{@{}l@{}}{\scriptsize  SRa: average success rate, \ SRwg:  success rate w/ grounding, \ SRog: success rate }\\[-1mm]
\multicolumn{8}{@{}l@{}}{\scriptsize  w/o grounding, Under.: understanding score, \ Appr.: appropriateness score.}
        \end{tabular}
    \label{tab:task1:human_eval}
\end{table}
\subsubsection{Summary}
Compared with the challenge results at DSTC8,  there is a  trend of shifting from building dialogs by assembling component-wise modules to end-to-end learning. In DSTC8, out of 11 teams with valid submissions,  1 team uses GPT-2 based models, 1 team uses word DST + word policy, with the rest 9 team uses component-wise models. This year, out of 10 teams, 8 teams used the end-to-end learning mechanism by leveraging transformer-based models. The top three systems in both automatic evaluation and human evaluation are all built using transformer-based end-to-end learning, and they have achieved much better performance in human evaluation than the systems at DSTC8\footnote{The success rate in DSTC8 human evaluation is success rate w/o grounding.}. 

\subsection{Cross-lingual Dialog State Tracking Task}
We introduce the task of cross-lingual dialog state tracking, requiring the participants to build a dialog state tracker for the target language with a training set in the source resource language and a small development set in the target language. 
Based on newly proposed large scale multi-domain task-oriented dialog datasets, MultiWOZ 2.1 \cite{eric2019multiwoz21} and CrossWOZ \cite{zhu2020crosswoz}, we offer two sub-tasks: 1) cross-lingual transfer from English to Chinese using MultiWOZ 2.1 dataset and 2) cross-lingual transfer from Chinese to English using CrossWOZ dataset. 

Following a similar scheme as in DSTC-5 \cite{kim2016fifth}, we provided machine translations of the original dataset.
We collected 500 new dialogs in the target language as the test set. 
The performance of each dialog state tracker is evaluated on the test set and compared with reference annotation.

\subsubsection{Data}
Compared with previous datasets \cite{kim2016fifth,multilingualwoz2017,schuster2019xlu} for cross-lingual transfer learning in task-oriented dialog, MultiWOZ 2.1 and CrossWOZ are much larger. MultiWOZ 2.1 contains over 10,000 dialogs, and CrossWOZ contains over 6,000 dialogs. They are also more challenging due to the multi-domain setting.
For each sub-task, we prepared data in a similar way:
a) collected 500 new dialogs in the source language, b) translated the ontology to the target language, and c) translated the original dialogs and the new dialogs. 
We released 250 new dialogs without any annotation as a public test set and reserved the other 250 dialogs as a private test set. 
\paragraph{Test Data Collection}
To collect new dialogs, we adapted the data collection website of CrossWOZ where paired workers can converse synchronously and make annotations.
New user goals were generated by the goal generator from ConvLab-2.
Following the Wizard-of-Oz setting, one worker acts as the user who needs to accomplish the allocated goal, and the other acts as the system that uses the database to provide information.
During the conversation, both sides need to annotate the dialog acts of their utterances, and the system should also annotate the dialog states that are queries over the database.

\paragraph{Ontology Translation}
We extracted the ontology from dialog act and dialog state annotations of both the original and test datasets.
Then we used Google Translate to translate them to the target language. 
For some slots that may not be faithfully translated, such as ``name'' and ``address'', we employed human translators to correct the translations.
This process is vital to ensure the translation consistency of the same values in different contexts.

\paragraph{Dialog Translation}
To make sure that the translations of values in a dialog are faithful to the ontology dictionary, we first replaced the values that appeared in the dialog with their translations in the dictionary.
Then we used Google Translate to translate the resulting code-switching sentences from the original dataset and test set.
In this way, translated dialogs and corresponding annotations do not conflict.
250 dialogs were sampled from the original dataset as the development set. Human translators were employed to proofread the translations of the development and test set.

\subsubsection{Evaluation Criteria}
\label{sec:task2:evaluation}
We evaluate the performance of the dialog state tracker using the following metrics:
a) Joint Goal Accuracy. This metric evaluates whether the predicted dialog state is exactly equal to the ground truth.
b) Slot Accuracy. This metric evaluates whether each slot's predicted label is exactly equal to the ground truth, averaged over all slots.
c) Slot Precision/Recall/F1. These metrics evaluate the overlap between the predicted labels and the ground truth for non-empty slots, micro-averaged over dialog turns.
Each submission contains the predictions for the public test set and the model that is used to make predictions for the private test set.
The results are averaged over the public and private test set.
The final ranking is solely based on the joint goal accuracy.

\subsubsection{Results}
The results of MultiWOZ (en$\rightarrow$zh) and CrossWOZ (zh$\rightarrow$en) sub-tasks are shown in Table \ref{tab:task2:multiwoz} and \ref{tab:task2:crosswoz_v1} respectively.
During the evaluation, we found that the CrossWOZ test data miss many ``name'' labels when the user accepts the attraction/hotel/restaurant recommended by the system.
Therefore, we utilized the database search results and heuristic rules to correct empty ``name'' labels and provided an updated leaderboard for CrossWOZ in Table \ref{tab:task2:crosswoz_v2}. Both of the CrossWOZ leaderboards are valid, but the updated one is preferred.

\newcolumntype{C}{>{\centering\arraybackslash}X}
\sisetup{table-column-width=11mm,round-precision=3,tight-spacing=false,table-format=1.3}
\begin{table}[H]
    \footnotesize
    \caption{MultiWOZ Leaderboard (Best Submissions).}
        \centering
        % \begin{tabularx}{\linewidth}{@{}Cccccccc@{}}
        % \setlength{\tabcolsep}{1.6mm}{
        \begin{tabular}{@{}cccccccc@{}}
        % \begin{tabularx}{\linewidth}{@{}cccccc@{}}
        \toprule
           Team & JGA & SA & Slot P/R/F1 & JGA(pub/pri) & Rank\\
        \midrule
1&   62.37&	98.09&	92.15/94.02/93.07&	62.70/62.03& 1  \\
2&   62.08&	98.10&	90.61/96.20/93.32&	63.25/60.91& 2  \\
3&   30.13&	94.40&	87.07/74.67/80.40&	30.53/29.72& 3  \\
BS&  55.56&	97.68&	92.02/91.10/91.56&	55.81/55.31& N/A \\
\bottomrule \\[-3.5mm]
\multicolumn{6}{@{}l@{}}{\scriptsize  JGA: joint goal accuracy, \ SA:  slot accuracy, \ Slot P/R/F1: slot precision/recall/f1,} \\[-1mm]
\multicolumn{6}{@{}l@{}}{\scriptsize  pub/pri: public/private test set.}
        % \end{tabularx}
        \end{tabular}
        % }
    \label{tab:task2:multiwoz}
\end{table}

\newcolumntype{C}{>{\centering\arraybackslash}X}
\sisetup{table-column-width=11mm,round-precision=3,tight-spacing=false,table-format=1.3}
\begin{table}[H]
    \footnotesize
    \caption{CrossWOZ Leaderboard (Best Submissions).}
        \centering
        % \begin{tabularx}{\linewidth}{@{}Cccccccc@{}}
        % \setlength{\tabcolsep}{1.6mm}{
        \begin{tabular}{@{}cccccccc@{}}
        % \begin{tabularx}{\linewidth}{@{}cccccc@{}}
        \toprule
           Team & JGA & SA & Slot P/R/F1 & JGA(pub/pri) & Rank\\
        \midrule
3&   16.86&	89.11&	68.26/62.85/65.45&	16.82/16.89& 1  \\
1&   15.28&	90.37&	65.94/78.87/71.82&	15.19/15.37& 2  \\
2&   13.99&	91.92&	72.63/78.90/75.64&	14.41/13.58& 3  \\
BS&  7.21&	85.13&	55.27/46.15/50.30&	7.41/7.00& N/A \\
\bottomrule \\[-3.5mm]
% \multicolumn{6}{@{}l@{}}{\scriptsize  JGA: joint goal accuracy, \ SA:  slot accuracy, \ Slot P/R/F1: slot precision/recall/f1,} \\[-1mm]
% \multicolumn{6}{@{}l@{}}{\scriptsize  pub/pri: public/private test set.}
        % \end{tabularx}
        \end{tabular}
        % }
    \label{tab:task2:crosswoz_v1}
\end{table}

\newcolumntype{C}{>{\centering\arraybackslash}X}
\sisetup{table-column-width=11mm,round-precision=3,tight-spacing=false,table-format=1.3}
\begin{table}[H]
    \footnotesize
    \caption{CrossWOZ Leaderboard (Updated Evaluation, Best Submissions).}
        \centering
        % \begin{tabularx}{\linewidth}{@{}Cccccccc@{}}
        % \setlength{\tabcolsep}{1.6mm}{
        \begin{tabular}{@{}cccccccc@{}}
        % \begin{tabularx}{\linewidth}{@{}cccccc@{}}
        \toprule
           Team & JGA & SA & Slot P/R/F1 & JGA(pub/pri) & Rank\\
        \midrule
2&   32.30&	94.35&	81.39/82.25/81.82&	32.70/31.89& 1  \\
1&   23.96&	92.94&	74.96/83.41/78.96&	23.45/24.47& 2  \\
3&   15.31&	89.70&	74.78/64.06/69.01&	14.25/16.37& 3  \\
BS&  13.02&	87.97&	67.18/52.18/58.74&	13.30/12.74& N/A \\
\bottomrule \\[-3.5mm]
% \multicolumn{6}{@{}l@{}}{\scriptsize  JGA: joint goal accuracy, \ SA:  slot accuracy, \ Slot P/R/F1: slot precision/recall/f1,} \\[-1mm]
% \multicolumn{6}{@{}l@{}}{\scriptsize  pub/pri: public/private test set.}
        % \end{tabularx}
        \end{tabular}
        % }
    \label{tab:task2:crosswoz_v2}
\end{table}

We adapted SUMBT \cite{lee2019sumbt} as the baseline model and used the translated training set of the original dataset to train for both sub-tasks.
We have received 10 models for MultiWOZ (en$\rightarrow$zh) and 8 models for CrossWOZ (zh$\rightarrow$en) from the same 3 teams. We briefly introduce their best models here.
Team 1 incorporated a four-class state operation prediction task into CHAN model \cite{chan2020dst}.
Team 2 modified SOM-DST \cite{kim2020somdst} and used ontology and some handcraft rules to post-process the generated values. 
Team 3 formulated the dialog state tracking as a sequence generation problem and used mBART to generate pairs of slot names and slot values. 
All of their best models were trained using the translated data in the target language.

\subsubsection{Summary}
To our surprise, all the best models are trained on monolingual machine translated data instead of both the original data and translations.
Team 2 and 3 even got negative results when training XLM/mBART on the original data and the translations simultaneously.
The performance of ``Translate-Train'' partially depends on the machine translator, which may be why team 1 and 2 augment the data by using another translator to translate the original dataset.
Team 1 and 2 modified DST models that are state-of-the-art on English MultiWOZ 2.1 dataset and got strong performance on Chinese MultiWOZ 2.1, verifying these models' language portability.
% Results on CrossWOZ are much lower than MultiWOZ,

\section{Track 3 - Interactive Evaluation of Dialog}

\subsection{Track overview}

The aim of dialog research is to create systems that can be effectively used in interactive settings by real users \cite{eskenazi2019beyond}. Despite this, the majority of research is performed on static datasets. For example, the task of response generation is typically done by producing a response for a static dialog context \cite{vinyals2015neural}. This track is intended to move dialog research beyond datasets and evaluate models in interactive environments with real users.

This track consists of two sub-tasks: (1) static evaluation and (2) interactive evaluation. The first subtask challenges participants to build response generation models which are evaluated in a static manner, using the Topical-Chat corpus \cite{gopalakrishnan2019topical}. The second subtask aims to extend dialog models beyond datasets and assess them in an interactive setting with real users, using DialPort \cite{zhao2016dialport}. In in the first subtask, models must generate a response to a fixed dialog context. In contrast in the second subtask, they must have a back-and-forth interaction with a real user. Through the two subtasks, this track challenges participants to take strong response generation models and develop strategies of making them effective in interactive settings. 

\subsection{Data}

Participants in this track were free to train on any publicly available data or use any pre-trained models. The static evaluation in the first subtask was carried out on the Topical-Chat corpus \cite{gopalakrishnan2019topical}. Topical-Chat is a large collection of human-human knowledge-grounded open-domain conversations that consists of 11,319 dialogs and 248,014 utterances. For each conversational turn, several relevant facts are provided. Models must leverage these facts and generate a response. This dataset was chosen because it is the largest, knowledge-grounded open-domain dataset presently available, to our knowledge. Additionally, the choice of usable facts provides a mechanism for systems to tailor responses to a specific user's interests. 

Since we continuously performed human evaluation over the duration of the challenge and used reference free evaluation metrics \cite{mehri2020usr}, it was not strictly necessary for models to be trained on the Topical-Chat corpus. A strong pre-trained dialog model may perform well on the first subtask, despite not training on the corpus.

The second subtask was not tied to a dataset. The interactive evaluation was carried out on DialPort\footnote{\url{http://dialog.speech.cs.cmu.edu:3000/}} \cite{zhao2016dialport} with real users recruited through Facebook Advertising. 

\subsection{Evaluation Criteria}

The first subtask was evaluated using ongoing (1) human evaluation and (2) three automatic metrics: METEOR \cite{banerjee2005meteor}, BERTscore \cite{zhang2019bertscore} and USR \cite{mehri2020usr}. Human evaluation was carried out on Amazon Mechanical Turk with the annotation questionnaire used to collect the FED dataset \cite{mehri2020unsupervised}. Over the duration of the challenge, we carried out evaluation on the Topical-Chat frequent validation set. For human evaluation, 30 context-response pairs were sampled and each one was labeled by 3 annotators. For the final evaluation, we carried out automatic evaluation on the frequent test set and perform human evaluation on 100 randomly sampled context-response pairs. For the final evaluation, the 100 dialog contexts used for evaluation were consistent across the different systems.

The evaluation for the second subtask consists of (1) collecting dialogs through conversations with real users on DialPort and (2) post-hoc assessment of the collected dialogs. Participants submitted dialog models (via an API) to DialPort. Real users were recruited through Facebook Advertising to interact with the submitted dialog systems. After gathering a sufficient number of conversations, we performed post-hoc assessment of the dialogs with the FED metric \cite{mehri2020unsupervised} and human evaluation on Amazon Mechanical Turk with the annotation questionnaire used to collect the FED dataset \cite{mehri2020unsupervised}.

Throughout the challenge, we aimed to collect at least \textit{100} conversations for each submitted system discounting any dialogs with offensive terms (e.g., curse words, racist phrases). For each system, 100 conversations were evaluated with the FED metric and on Amazon Mechanical Turk, with 3 annotators labeling each dialog. 

For the final submission, we gather dialogs for all systems over the same time period. Ultimately, given a Facebook Advertising budget of \$2500 and 11 systems (including two baselines), we obtained 4651 conversations (after removing offensive dialogs) with a total of 41,640 turns. We consider only the conversations that are at least four turns in length (total of 2960) for the final post-hoc assessment. For each system, we carry out human evaluation with 200 conversations of suitable length. Throughout the challenge, all individuals who interact with the system on DialPort \textit{do so for free, of their own volition}, thereby avoiding common problems observed with paid users \cite{ai2007comparing}. 

\subsection{Results}

The challenge received \textbf{33} submissions to the first subtask and \textbf{9} submissions to the second subtask.

\newcolumntype{C}{>{\centering\arraybackslash}X}
\sisetup{table-column-width=11mm,round-precision=3,tight-spacing=false,table-format=1.3}
\begin{table}
    \footnotesize
    \caption{Results for subtask 1. For brevity, we only show the top 10 submissions (out of 33) according to the human evaluation. This table only reports the overall USR metric and the overall impression of the response from the human evaluation. The full evaluation results may be found     \href{https://docs.google.com/spreadsheets/d/1FWRUA1MFwe0IWFpHnrVr6Pwo6VGU6gjLYNPHrq5Qs4w/}{here.}}
        \centering
        % \begin{tabularx}{\linewidth}{@{}Cccccccc@{}}
        \begin{tabular}{@{}cccccc@{}}
        \toprule
           System & METEOR & BERTscore & USR & Human & Rank \\
        \midrule
     
1  &     9.06 &   84.91 &  4.26 &     \textbf{4.281} & 1 \\
2  &     13.11 &   86.17 &  4.59 &    \textbf{4.280} & 1\\
3  &     6.83 &   84.36 &  3.86 &      \textbf{4.280} & 1\\
4  &     8.96 &   85.15 &  4.26  &     4.260 & 4\\
5  &     12.37 &   86.21 &  \textbf{4.83} &    4.253 & 5\\
6  &     12.31 &   86.32 &  4.73 &     4.231 & 6\\
7  &     \textbf{13.96} &   \textbf{86.84} &  4.48 &     4.229 & 6\\
8  &     12.51 &   85.91 &  4.45 &      4.229 & 6\\
9  &     12.14 &   85.91 &    4.46 & 4.216 & 9\\
10 &     10.87 &   85.65 &  4.53 &     4.210 & 10\\
\bottomrule \\[-3.5mm]
        \end{tabular}
    \label{tab:track3_task1}
\end{table}

Table \ref{tab:track3_task1} shows the results of the static evaluation on the Topical-Chat corpus \cite{gopalakrishnan2019topical}, for the 10 best performing systems according to the human evaluation. All of the top 10 systems used either pre-trained models or additional data, highlighting the importance of pre-training for open-domain response generation. This observation aligns with previous research, which has seen strong performance in open-domain response generation through the use of large-scale pre-training \cite{zhang2019dialogpt, adiwardana2020towards}. 

In addition to performing ongoing human evaluation throughout the challenge, we assess systems in the first subtask using three evaluation metrics. METEOR \cite{banerjee2005meteor} and BERTscore \cite{zhang2019bertscore}, are referenced evaluation metrics that compare a generated output to a \textit{ground-truth response}. In contrast, USR \cite{mehri2020usr} is a reference free evaluation metric that uses pre-trained models and self-supervised training objectives to estimate the quality of a response. Though none of the evaluation metrics is a perfect predictor of the final ranking, we find that USR better correlates with the system-level human performance (Spearman: 0.35, $p < 0.05$) than either METEOR (Spearman: 0.23, $p > 0.05$) or BERTscore (Spearman: 0.22, $p > 0.05$). The relatively low system-level correlation highlights the importance of performing ongoing human evaluation throughout the challenge. 

The poor performance of automatic metrics, may in part be a consequence of the fact that several submissions did not fine-tune on the Topical-Chat corpus and instead relied on open-domain response generation capabilities learned through large-scale pre-training. As such, while the responses were favored by human annotators - the automatic metrics penalized them for either not having high word-overlap with the ground truth (METEOR, BERTscore) or not resembling the utterances in the Topical-Chat corpus (USR).

\newcolumntype{C}{>{\centering\arraybackslash}X}
\sisetup{table-column-width=11mm,round-precision=3,tight-spacing=false,table-format=1.3}
\begin{table}
    \footnotesize
    \caption{Results for subtask 2. This table reports for each system: the overall FED metric, the overall impression of the dialogs from the human evaluation, as well as the average number of dialog turns. The full results be found     \href{https://docs.google.com/spreadsheets/d/1FWRUA1MFwe0IWFpHnrVr6Pwo6VGU6gjLYNPHrq5Qs4w/edit\#gid=1829761446}{here.} System 6 and 11 are our DialoGPT and Transformer baselines, respectively.}
        \centering
        % \begin{tabularx}{\linewidth}{@{}Cccccccc@{}}
        \begin{tabular}{@{}ccccc@{}}
        \toprule
           System & Avg. Turns & FED & Human & Rank \\
           \midrule
    1  & 12.44& \textbf{4.97}    & \textbf{4.15}& 1  \\
    2  & \textbf{13.47}& 4.79    & 4.14& 2 \\
    3  & 8.89 & 4.61    & 4.08& 3 \\
    4  & 9.36 & 4.68    & 4.03& 4 \\
    5  & 9.82 & 4.53    & 3.93& 5 \\
    6 & 8.75 & 4.72  & 3.87& 6 \\
    7  & 8.51 & 4.41    & 3.85& 7 \\
    8  & 7.67 & 4.30     & 3.85& 7 \\
    9  & 6.53 & 4.64    & 3.83 & 9 \\
    10  & 7.35 & 4.80    & 3.69& 10 \\
    11  & 5.80 & 3.69        & 3.60 &  11 \\
\bottomrule \\[-3.5mm]
        \end{tabular}
    \label{tab:track3_task2}
\end{table}

The results for the second subtask are shown in Table \ref{tab:track3_task2}. System 6 is our DialoGPT baseline \cite{zhang2019dialogpt}, fine-tuned on the Topical-Chat corpus without knowledge grounding. System 11 is our Transformer baseline which was trained on the Topical-Chat corpus and uses tf-idf sentence similarity to retrieve relevant knowledge at inference time. The best performing model, \textbf{System 1}, leverages large-scale pre-training in addition to strategies for producing more diverse responses. This system achieved first place in both subtasks: \textbf{System 1} in Table \ref{tab:track3_task2} corresponds to \textbf{System 2} in Table \ref{tab:track3_task1}. 

FED \cite{mehri2020unsupervised}, which is an \textit{unsupervised} evaluation metric for interactive dialog is shown to be a moderate predictor of the final ranking with a system-level Spearman correlation of 0.49 ($p = 0.13$), though it correctly predicts the top two systems. We also note that the average number of turns for a particular system is a strong indicator of its quality here (Spearman: 0.94, $p < 0.01$). Real users are more inclined to interact with a better system, making it an important metric for assessing systems in interactive settings \cite{ram2018conversational}.  

While many of the submissions in the first subtask perform similarly, the scores in Table \ref{tab:track3_task2} are much more varied. This signifies that interactive evaluation more exhaustively tests the capabilities of systems and is therefore a more indicative measure of a system's capabilities. This observation has been shown by prior work \cite{mehri2020unsupervised}, when analyzing dialogs from Meena \cite{adiwardana2020towards}.

The \textit{Interactive Evaluation of Dialog} track demonstrates both the feasibility and the importance of evaluating dialog systems in interactive settings with real users. We show that with an advertising budget of \$2500, we collect more than 4000 dialogs on DialPort (2960 dialogs with at least 4 turns). The results of interactive evaluation are more varied (Table \ref{tab:track3_task2}) suggesting that back-and-forth interactions with real users are challenging to dialog systems and that interactive evaluation is a better reflection of a system's capabilities.
\section{Track 4 - SIMMC: Situated Interactive Multi-Modal Conversational AI}

\subsection{Track overview}

The SIMMC challenge aims to lay the foundations for the real-world assistant agents that can handle multimodal inputs, and perform multimodal actions. We thus focus on task-oriented dialogs that encompass a situated multimodal user context in the form of a co-observed image or virtual reality (VR) environment. The context is dynamically updated on each turn based on the user input and the assistant action.
%Our challenge focuses on our SIMMC datasets, both of which are shopping domains: (a) furniture (grounded in a shared virtual environment) and, (b) fashion (grounded in an evolving set of images).
Moon {\it et al.}~\cite{moon2020situated} provides more details on the datasets and the models we provide.

\subsection{Data}

\begin{table*}[t]
    \begin{center}
        \caption{
        Summary of each team's results on Test-Std split, average of Furniture and Fashion (*Team 5 submitted results only for Fashion).
        Best results from each team are shown.
        \textbf{(1) API prediction} via \underline{acc}uracy, \underline{perp}lexity and \underline{a}ttribute \underline{acc}uracy, and,
        \textbf{(2) Response prediction} via \underline{BLEU}, 
        \underline{r}ecall\underline{@k} (k=1,5,10), \underline{mean} rank, and mean
        reciprocal rank (\underline{MRR}).
        \textbf{(3) Dialog State Tracking (DST)}, via \underline{slot} and \underline{intent} prediction \underline{F1}.
        $\uparrow$: higher is better, $\downarrow$: lower is better.
        }
    \label{tab:simmc:results}
        \begin{tabular}{
            p{0.13\columnwidth}
            cccccccccccc
        }
        \toprule[\heavyrulewidth]
        \multirow{2}{*}{\textbf{Teams}}
        & \multicolumn{3}{c}{\textbf{Subtask 1. API Prediction}}
        & \multicolumn{6}{c}{\textbf{Subtask 2. Response Generation}}
        & \multicolumn{2}{c}{\textbf{Subtask 3. DST}} \\
        \cmidrule(r){2-4}
        \cmidrule(r){5-10}
        \cmidrule(r){11-12}
            & Acc$\uparrow$ 
            & Perp$\downarrow$
            & A.Acc$\uparrow$
            & BLEU$\uparrow$ 
            & r@1$\uparrow$ 
            & r@5$\uparrow$
            & r@10$\uparrow$
            & Mean$\downarrow$
            & MRR$\uparrow$
            & Slot F1$\uparrow$
            & Intent F1$\uparrow$\\
        \midrule
        Baseline
            & \reportvalbrief{79.3}{0}
            & \reportvalbrief{63.7}{0}
            & \reportvalbrief{1.9}{0}
            & \reportvalbrief{0.0061}{0}
            & \reportvalbrief{0.145}{0}
            & \reportvalbrief{7.2}{0}
            & \reportvalbrief{19.8}{0}
            & \reportvalbrief{27.3}{0}
            & \reportvalbrief{39.2}{0}
            & \reportvalbrief{62.4}{0}
            & \reportvalbrief{62.1}{0}\\
        Team 1
            & \reportvalbrief{80.2}{0}
            & \reportvalbrief{74.6}{0}
            & \reportvalbrief{2.0}{0}
            & \reportvalbrief{0.105}{0}
            & \reportvalbrief{0.326}{0}
            & \reportvalbrief{21.1}{0}
            & \reportvalbrief{43.6}{0}
            & \reportvalbrief{56.8}{0}
            & \reportvalbrief{18.8}{0}
            & \reportvalbrief{77.8}{0}
            & \reportvalbrief{76.7}{0}\\            
        Team 2
            & \reportvalbrief{\bf 82.5}{0}
            & \reportvalbrief{69.8}{0}
            & \reportvalbrief{1.8}{0}
            & \reportvalbrief{0.082}{0}
            & \reportvalbrief{0.074}{0}
            & \reportvalbrief{2.5}{0}
            & \reportvalbrief{8.3}{0}
            & \reportvalbrief{13.6}{0}
            & \reportvalbrief{47.7}{0}
            & -
            & -\\            
        Team 3
            & \reportvalbrief{79.4}{0}
            & \reportvalbrief{73.2}{0}
            & -
            & \reportvalbrief{\bf 0.128}{0}
            & \reportvalbrief{0.381}{0}
            & \reportvalbrief{26.3}{0}
            & \reportvalbrief{50.3}{0}
            & \reportvalbrief{61.8}{0}
            & \reportvalbrief{15.5}{0}
            & \reportvalbrief{\bf 79.1}{0}
            & \reportvalbrief{78.1}{0}\\            
        Team 4
            & \reportvalbrief{\bf 81.3}{0}
            & \reportvalbrief{\bf 73.9}{0}
            & \reportvalbrief{3.5}{0}
            & \reportvalbrief{0.108}{0}
            & \reportvalbrief{\bf 0.673}{0}
            & \reportvalbrief{52.6}{0}
            & \reportvalbrief{87.4}{0}
            & \reportvalbrief{95.1}{0}
            & \reportvalbrief{3.2}{0}
            & \reportvalbrief{78.6}{0}
            & \reportvalbrief{77.7}{0}\\            
        Team 5*
            & -
            & -
            & -
            & -
            & \reportvalbrief{0.39}{0}
            & \reportvalbrief{26.7}{0}
            & \reportvalbrief{52.1}{0}
            & \reportvalbrief{66.0}{0}
            & \reportvalbrief{14.8}{0}
            & -
            & -\\                 
        \bottomrule[\heavyrulewidth]
        \end{tabular}
    \end{center}
    %\vspace*{-10pt}
\end{table*}
% Results taken from https://fb.workplace.com/notes/satwik-kottur/simmc-challenge-result-announcement-in-dstc9/731516807433850/

SIMMC contains about $13k$ human-to-human dialogs (totaling about $169k$ utterances). We chose shopping experiences---specifically furniture and fashion---as the domain for the SIMMC datasets because of the dynamic environment created by these domains, where rich multimodal interactions happen around visually grounded items.

SIMMC offers four key advantages over previous multimodal dialog datasets:
\begin{enumerate}
    \item SIMMC assumes a co-observed multimodal context between a user and an assistant and records the ground-truth item appearance logs of each item that appears. SIMMC tasks emphasize semantic processing of the input modalities, while work in this area has traditionally focused heavily on raw image processing.
    \item Compared with the conventional task-oriented conversational datasets, the agent actions in the SIMMC datasets span across a diverse multimodal action space (e.g. ``rotate'', ``search'', and ``add to cart'').
    \item Agent actions can be enacted on both the object level (e.g. changing the view of a specific object within a scene) and the scene level (e.g. introducing a new scene or an image).
    \item SIMMC emphasizes semantic processing. The proposed SIMMC annotation schema allows for a more systematic and structural approach for visual grounding of conversations, which is essential for solving challenging problems in real-world scenarios.
\end{enumerate}
Datasets were collected through the SIMMC Platform \cite{crook2019simmc}, an extension to ParlAI \cite{miller2017parlai} for multimodal conversational data collection and system evaluation that allows human annotators to each play the role of either the assistant or the user.

\subsection{Evaluation Criteria}

We present three subtasks primarily aimed at replicating human-assistant actions in order to enable rich and interactive shopping scenarios.

\noindent \textbf{Subtask 1: Structural API Call Prediction} focuses on predicting the assistant action as an API call given the dialog and the multimodal contexts as inputs.
Since accuracy does not account for the existence of multiple valid actions, 
we use perplexity (defined as the exponentiation of the Shannon entropy) alongside accuracy. 
To also measure the correctness of the predicted action (API) arguments, we use attribute accuracy
compared to the collected datasets.

\noindent
\textbf{Subtask 2: Response Prediction} examines the relevance of the assistant response in the current turn.
We evaluate in two ways; \emph{(a)} as a conditional language modeling problem, where the closeness between the generated and
ground-truth response is measured through using BLEU-4 score, and,
\emph{(b)} as a retrieval problem, where we measure the model performance when retrieving ground-truth responses from
a pool of $100$ candidates (randomly chosen and unique to each turn).

\noindent
\textbf{Subtask 3: Dialog State Tracking (DST)} aims to systematically track the dialog acts and the associated slot pairs across multiple turns, as represented in the flexible ontology developed to represent the SIMMC multimodal context.
We use the intent and slot prediction metrics (F1), inline with prior work in DST.

\subsection{Results}

The challenge saw a total of 11 model entries from 5 teams across the world, setting a new state-of-the-art in all three subtasks (Table~\ref{tab:simmc:results}). 

For each subtask, we listed metrics in a priority order and the entry with the most favorable performance on the highest priority metric was considered to be a candidate winner. 
Any entries within one standard error of this candidate’s performance were also considered as candidates. 
Where there were more than one candidate, as in subtask 1, we used the next metric in the priority list and repeat this process until we had a single winner.

\textbf{The winner of the structural API call prediction subtask (subtask 1)} was a BART \cite{lewis-etal-2020-bart} model  (BART-Large) from Team 4 that jointly predicted the dialog state (subtask 3), API call (subtask 1) and response (subtask 2a) as single target given the dialog history, multimodal context and user utterance. This model was one of two runners up on subtask 2a, and the runner up on subtask 3.

\textbf{The winner of the response retrieval subtask (subtask 2b)} was a BART-based Bi-encoder \cite{DBLP:conf/iclr/HumeauSLW20, mazare-etal-2018-training, dinan2019wizard}, also from Team 4, whose weights were initialized from the jointly trained BART model that won subtask 1.
% Its weights were then adapted to the retrieval task.
This model achieved a mean reciprocal rank (MRR) of $0.67$, a lead of $0.29$ points ahead of the runner up team on this subtask. 

\textbf{The winner of the response generation and DST subtasks (subtask 2a and subtask 3)} was an ensemble of GPT-2 \cite{radford2019language} models from Team 3 that were of differing sizes (large and small) and used differing portions of the training and development sets. Each GPT-2 model was independently trained on the joint tasks---subtask 2a and subtask 3---using a simple language model loss that optimized over the concatenated dialog history, multimodal context, user utterance, dialog state and response. Preprocessing over dialogue states was done before training, and an ensemble beam search over each model’s prediction was used to generate the final prediction.

%\subsection{Summary}

\section{Conclusions}
This paper summarizes the four tracks of the ninth dialog system technology challenges (DSTC9).
Beyond Domain APIs track expands the coverage of current task-oriented  dialog systems  by incorporating  external unstructured knowledge sources.
Multi-domain Task-oriented Dialog Challenge II, focuses  on  end-to-end  multi-domain  task completion dialog and cross-lingual multi-domain dialog state tracking. Interactive  Evaluation  of Dialog Track, expands dialog research beyond  datasets  encourages to develop dialog  systems  that  can  converse  effectively  in  interactive environments. The Situated  Interactive Multi-Modal Conversational AI track focuses on real-world assistant agents that can handle multi-modal inputs, and perform multi-modal actions. All the datasets and resources introduced for every track will be publicly available even after the challenge period to support future dialog system research.

\bibliographystyle{IEEEtran}
\bibliography{main}

\end{document}